\documentclass[journal]{IEEEtran}

\pdfoutput=1

\usepackage[T1]{fontenc} 
\usepackage[utf8]{inputenc}
\usepackage[pdftex]{graphicx}
\usepackage[hidelinks]{hyperref}
\usepackage{cite}
\usepackage{siunitx}
\usepackage[super]{nth}
\usepackage{csquotes}
\usepackage{cleveref}
\usepackage{booktabs}
\usepackage{makecell}
\usepackage[usenames]{xcolor}
\usepackage[normalem]{ulem}
\usepackage[percent]{overpic}

\newcommand{\CNN}[1]{CNN\textsubscript{#1}}
\renewcommand{\vec}[1]{\mathbf{#1}}
\newcommand{\quartercaption}[1]{\makebox[0.24\textwidth]{\footnotesize #1}}
\newcommand{\onimagecaption}[1]{\color{white}\scriptsize\textsf{#1}}

\makeatletter
\newcommand*{\rom}[1]{\expandafter\@slowromancap\romannumeral #1@}
\makeatother

\newcommand{\midruleheading}[2]{%
	\midrule%
	\addlinespace[0.45em]%
	\multicolumn{#1}{l}{\emph{#2}}\\
	\addlinespace[0.25em]%
}

\IEEEpubid{\href{https://dx.doi.org/10.1109/TMI.2017.2769839}{\textbf{Published in IEEE Transactions on Medical Imaging 37(2), 615--625, 2018.}}}

\begin{document}

\title{Automatic calcium scoring in low-dose chest CT using~deep~neural~networks~with~dilated~convolutions}

\author{
	Nikolas~Lessmann,~Bram~van~Ginneken,~Majd~Zreik,~Pim~A.~de~Jong,~Bob~D.~de~Vos,\\Max~A.~Viergever,~Ivana~I\v{s}gum%
	\thanks{N.~Lessmann, M.~Zreik, B.\,D.~de Vos, M.\,A.~Viergever and I.~I\v{s}gum are with the Image Sciences Institute, University Medical Center Utrecht, Utrecht University, The Netherlands (e-mail: n.lessmann@umcutrecht.nl).}%
	\thanks{B. van Ginneken is with the Diagnostic Image Analysis Group at the Department of Radiology and Nuclear Medicine, Radboud University Medical Center, Nijmegen, The Netherlands, and Fraunhofer MEVIS, Bremen, Germany.}%
	\thanks{P.\,A. de Jong is with the Department of Radiology, University Medical Center Utrecht, Utrecht University, The Netherlands.}%
	\thanks{Copyright (c) 2017 IEEE. Personal use of this material is permitted. However, permission to use this material for any other purposes must be obtained from the IEEE by sending a request to pubs-permissions@ieee.org.}
}

\markboth{}{}
\maketitle


\begin{abstract}
	Heavy smokers undergoing screening with low-dose chest CT are affected by cardiovascular disease as much as by lung cancer. Low-dose chest CT scans acquired in screening enable quantification of atherosclerotic calcifications and thus enable identification of subjects at increased cardiovascular risk.
	This paper presents a method for automatic detection of coronary artery, thoracic aorta and cardiac valve calcifications in low-dose chest CT using two consecutive convolutional neural networks. The first network identifies and labels potential calcifications according to their anatomical location and the second network identifies true calcifications among the detected candidates. This method was trained and evaluated on a set of \num{1744} CT scans from the National Lung Screening Trial. To determine whether any reconstruction or only images reconstructed with soft tissue filters can be used for calcification detection, we evaluated the method on soft and medium/sharp filter reconstructions separately. On soft filter reconstructions, the method achieved F$_1$ scores of \num{0.89}, \num{0.89}, \num{0.67}, and \num{0.55} for coronary artery, thoracic aorta, aortic valve and mitral valve calcifications, respectively. On sharp filter reconstructions, the F$_1$ scores were \num{0.84}, \num{0.81}, \num{0.64}, and \num{0.66}, respectively. Linearly weighted kappa coefficients for risk category assignment based on per subject coronary artery calcium were \num{0.91} and \num{0.90} for soft and sharp filter reconstructions, respectively. These results demonstrate that the presented method enables reliable automatic cardiovascular risk assessment in all low-dose chest CT scans acquired for lung cancer screening.
\end{abstract}

\IEEEpeerreviewmaketitle


\section{Introduction}

\IEEEPARstart{S}{creening} with low-dose chest CT has been found effective in reducing mortality from lung cancer in current or former heavy smokers\cite{NLST2011}. However, smoking is not only a major risk factor for lung cancer, but also for cardiovascular disease (CVD)\cite{Vollset2006}. The presence of CVD can be detected in CT scans by measuring the amount of coronary artery calcification (CAC), a strong and independent predictor of cardiovascular events. CAC is usually quantified in dedicated cardiac CT images and expressed as a calcium score\cite{Hecht2015}. Recent studies have shown that calcium scores are also able to predict cardiovascular events if quantified in low-dose chest CT for lung cancer screening\cite{Shemesh2010,Jacobs2012,Chiles2015}. Calcium scoring could therefore complement lung cancer screening programs to help identify subjects at elevated cardiovascular risk without the need for further imaging \cite{Mets2013,Hecht2014}. However, manual calcium scoring in addition to lung screening would impose a considerable extra burden on screening programs due to the large number of subjects, the high average calcium burden in a high risk population, the suboptimal image quality of low-dose screening scans, and the lack of ECG synchronization leading to cardiac motion artifacts. Automatic calcium scoring could therefore be a viable alternative that would enable routine cardiovascular risk prediction from low-dose chest CT scans.

\IEEEpubidadjcol

Automatic methods for CAC scoring have been developed mostly for dedicated non-contrast enhanced cardiac CT \cite{Isgum2007,Kurkure2010,Brunner2010,Wu2012,Ding2015,Shahzad2013,Wolterink2015} or cardiac CT angiography (CTA) scans \cite{Dey2009,Wesarg2006,Eilot2014,Ahmed2015,Mittal2010,Wolterink2015b,Wolterink2016}.
Only few methods have been developed specifically for coronary calcium scoring in chest CT \cite{Isgum2012,Xie2014,Lessmann2016,Gonzalez2016}.

In non-contrast cardiac CT scans, the coronary arteries are visible only when calcified or embedded in fat. Automatic scoring methods therefore typically rely on segmentation or rough localization of larger structures such as the heart and the aorta to derive a region of interest \cite{Wu2012,Ding2015} or to derive spatial features for classification of candidate lesions using machine learning\cite{Isgum2007,Kurkure2010,Brunner2010}. Other methods that use machine learning derive spatial features from segmentations of the coronary arteries, which are obtained by registration of the non-contrast scan with a CTA-based atlas\cite{Shahzad2013,Wolterink2015}. Commonly used features besides spatial features are texture, lesion volume and shape. Spatial features are consistently reported to be most important.

In cardiac CTA scans, the coronary arteries are well visible due to the arterial contrast enhancement. Automatic scoring methods therefore typically perform a segmentation of the coronary artery tree. The segmentation is used to detect calcifications by searching for strong intensity gradients along the segmented vessel \cite{Dey2009,Wesarg2006,Eilot2014,Ahmed2015} because calcifications are typically brighter than the contrast enhanced lumen of the vessel. In a similar approach, Mittal~et~al.\cite{Mittal2010} employ a classifier to detect calcifications based on features that describe the texture along the vessel. In contrast to such approaches, Wolterink~et~al.\cite{Wolterink2015} proposed a method without prior segmentation of the coronary arteries. The method uses a convolutional neural network (CNN) in combination with simple spatial features based on image coordinates to classify candidate voxels in the image. To reduce the false positive rate of the CNN, connected groups of voxels that were detected by the CNN are reclassified by a random forest classifier. In their later publication \cite{Wolterink2016}, a second CNN replaces this step.

In non-contrast chest CT scans, segmentation of the coronary artery tree is not feasible due to the lack of contrast enhancement and due to cardiac motion artifacts caused by the lack of ECG synchronization. Automatic scoring methods therefore typically rely on other means to identify a region of interest, similar to methods for CAC detection in non-contrast cardiac CT. Išgum~et~al.\cite{Isgum2012} obtain spatial features from a coronary calcium probability atlas and use these in combination with volume and texture features in a multi-classifier approach. Xie~et~al.\cite{Xie2014} and González~et~al.\cite{Gonzalez2016} segment or roughly localize the heart and identify coronary calcifications in the detected region of interest based on decision rules. In our preliminary work, we proposed to use a CNN to classify candidate voxels within a bounding box around the heart \cite{Lessmann2016}.

In addition to coronary calcifications, thoracic aorta calcifications (TAC) and calcifications of the cardiac valves have been related to cardiovascular risk\cite{Tison2015,Jacobs2010,Willemink2015}.
For calcium scoring in the thoracic aorta in chest CT, few automatic methods have been published. All methods first perform a segmentation of the aorta followed by either rule-based calcification detection using auxiliary segmentations of trachea and spine \cite{Kurugol2012,Xie2014b} or calcification detection based on machine learning \cite{Isgum2010}. The machine learning approach uses kNN classifiers with features similar to those used by methods for coronary calcium detection: various spatial features derived from the segmentation of the aorta, volume of the potential calcification and texture features.
For detection and quantification of cardiac valve calcifications, automatic methods have not been published.

We propose an automatic system for concurrent detection of CAC, TAC and cardiac valve calcifications in low-dose chest CT. These calcifications likely show different aspects of atherosclerotic disease and their quantities can potentially complement each other in detecting the presence of CVD and in predicting cardiovascular events. In contrast to simple combination of the output of multiple systems, e.g., one system for CAC and another for TAC detection, a single method for concurrent detection avoids ambiguous results.

Furthermore, we propose to label each voxel separately according to the affected vessel rather than labeling calcified lesions, i.e., connected voxels above a certain intensity value. Lesion labeling is standardly performed in clinically used commercial software and most previous automatic methods. However, voxel labeling allows separation of single lesions that extend to multiple vascular beds, e.g., in the aorta and a coronary artery. This is important because calcifications in different arteries carry different prognostic value \cite{Jacobs2010}.

Next, we propose to use a CNN to directly identify potential calcifications within the entire image, without the need to explicitly segment or localize anatomical structures.
The majority of methods in the literature for CAC or TAC detection rely on segmentation to restrict the region of interest or to infer spatial features as these have been found important for accurate calcium detection.
Instead, context information is not provided by segmentations, but rather the CNN has to be able to infer context information directly from the image. We therefore use a CNN with a particularly large receptive field, which is achieved by using an architecture based on dilated convolutions. This large receptive field furthermore enables the network to label candidates based on their anatomical location. Similar to \cite{Wolterink2016}, we subsequently employ a second CNN to identify calcifications among the candidates identified and labeled by the first CNN. The main contribution is therefore a network architecture tailored specifically to the problem of calcium detection in low-dose chest CT scans.

Finally, we evaluated our method on a large and diverse dataset from the to date largest lung cancer screening trial with low-dose chest CT. Most other methods for automatic CAC or TAC scoring have been evaluated only on relatively small and homogeneous datasets. However, a method that is applied clinically will face images acquired with a multitude of scanner models and reconstructed with a wide range of reconstruction algorithms. In this work, scans were therefore selected such that a wide variety of screening sites, scanner models and reconstruction algorithms are present.

\section{Dataset} \label{sec:Data}

\begin{table}
	\centering
	\caption{Overview of scanner models and reconstruction filters}
	\label{tbl:ScannerModels}
	\begin{tabular}{lll}
		\toprule
		Vendor & Models & Reconstruction filters\\
		\midrule
		General Electric & LightSpeed 16 & Standard (soft tissue) \\
		(GE) & LightSpeed Pro 16 & Bone, Lung (medium/sharp) \\
		& LightSpeed Ultra &  \\
		& LightSpeed Plus & \\
		& LightSpeed QX/i & \\		
		& Discovery QX/i & \\
		& HiSpeed QX/i & \\
		\addlinespace[0.3em]
		Siemens & Sensation 4 & B30f (soft tissue)\\
		& Sensation 16 & B50f, B80f (medium/sharp) \\
		& Volume Zoom & \\
		\addlinespace[0.3em]
		Philips & MX8000 & C (soft tissue)\\
		& MX8000 IDT & D (medium/sharp) \\
		\addlinespace[0.3em]
		Toshiba & Aquilion & FC10 (soft tissue)\\
		& & FC51 (medium/sharp)\\
		\bottomrule
	\end{tabular}
\end{table}

\begin{figure}
	\includegraphics[width=0.155\textwidth]{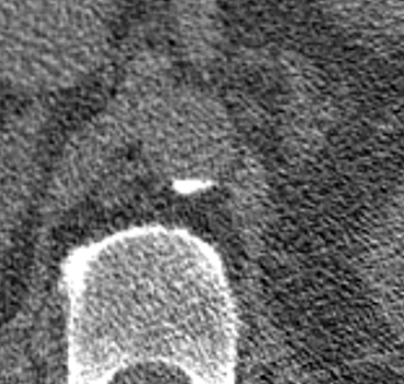}%
	\hfill%
	\includegraphics[width=0.155\textwidth]{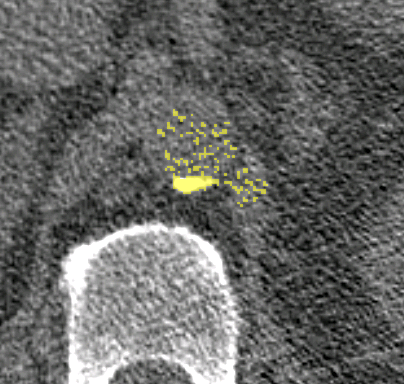}%
	\hfill%
	\includegraphics[width=0.155\textwidth]{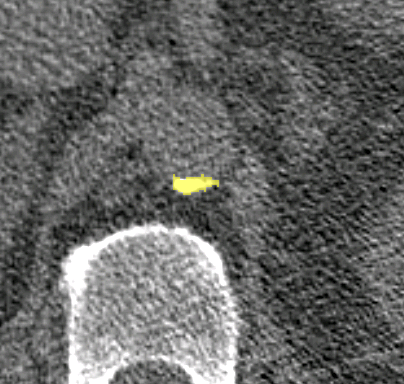}%
	\caption{Segmentation of a TAC lesion in a noisy scan (left) using 3D region growing (center) and fully manual segmentation (right). Region growing here leads to segmentation of a large amount of noise together with the calcium.}
	\label{fig:RegionGrowingNoise}
\end{figure}

\begin{figure*}
	\centering
	\includegraphics[width=0.8\textwidth]{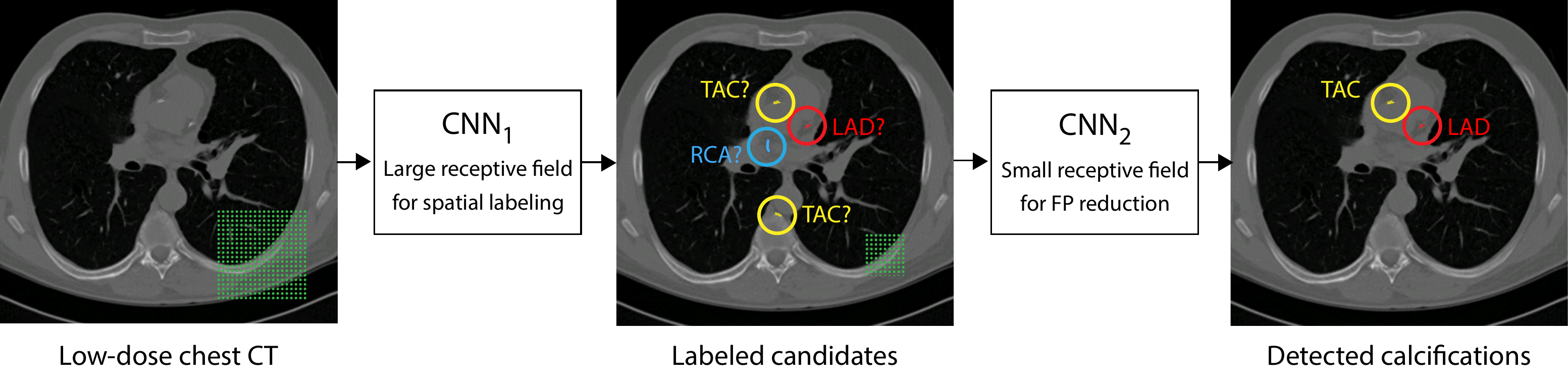}
	\caption{Overview of the proposed calcium detection method. Two CNNs are applied consecutively to first identify and label candidates (\CNN{1}), and to finally identify true calcifications among the candidates (\CNN{2}). The size of the receptive fields of the networks are indicated by green dotted areas.}
	\label{fig:Flowchart}
\end{figure*}

For training and evaluation, we used low-dose chest CT scans acquired in the National Lung Screening Trial (NLST). The NLST was a large lung cancer screening trial in the United States that enrolled \num{53454} current or former heavy smokers aged 55 to 74\cite{NLST2011}. To develop and evaluate the proposed automatic calcium scoring method on a diverse data set, we selected \num{1744} scans from \num{6000} available baseline scans by randomly sampling from the scans acquired with the 25 most common imaging settings with respect to scanner model and reconstruction algorithm. Specifically, 100 scans were selected for each of the 10 most common settings, and up to 50 for each of the 15 next most common settings.

The selected scans were acquired in 31 medical centers on 13 different scanner models from four major CT scanner vendors (\Cref{tbl:ScannerModels}). The scans were acquired at breath-hold after inspiration in helical scanning mode without contrast enhancement and without ECG synchronization. Tube voltage was set to 120 kVp, or 140 kVp for large subjects (\SI{5}{\percent}). In-plane resolution ranged from \SI{0.49}{\milli\meter} to \SI{0.98}{\milli\meter}, slice thickness from \SI{1}{\milli\meter} to \SI{3}{\milli\meter} and slice spacing from \SI{0.6}{\milli\meter} to \SI{3.0}{\milli\meter}. Since calcium scoring is typically performed on \SI{3}{\milli\meter} thick slices, we reconstructed \SI{3}{\milli\meter} thick axial slices with \SI{1.5}{\milli\meter} slice spacing from all scans.

To establish a reference standard, calcifications were manually labeled in all scans. Scans were distributed among four trained observers and one radiologist with extensive experience in calcium scoring. To measure interobserver agreement, a subset of 100 scans (four scans from each of the 25 different scanner models and reconstruction algorithms) was annotated by two of the trained observers and the radiologist. Manual calcium annotation usually requires the observer to select only a single voxel per lesion. The lesion is then automatically segmented with region growing using the standard intensity threshold of \SI{130}{HU}. In low-dose scans, however, intensity based region growing often leads to large amounts of noise being segmented with the calcium (\Cref{fig:RegionGrowingNoise}). Moreover, it can lead to the spine and ribs being segmented together with calcium, or calcifications in arteries branching off the aorta being segmented together with calcium in the aorta. The observers therefore marked calcifications voxel-by-voxel ($\ge$\SI{130}{HU}) in the coronary arteries, the aorta and the aortic and mitral valves, including the annulus. Coronary calcifications were labeled as either left anterior descending artery (LAD), left circumflex artery (LCX) or right coronary artery (RCA). The left main coronary artery was considered part of LAD because these are difficult to distinguish on ungated scans. Motion artifacts caused by calcifications were annotated as calcifications because an exact separation of true calcification and artifact is often not possible. Depending on the amount of calcification and the image quality, the annotation effort varied from 5--10 minutes for images with soft reconstruction and little calcium to 60--90 minutes for images with sharp reconstruction and/or large amounts of calcium.

\section{Method} \label{sec:Method}

The proposed method for automatic detection of CAC, TAC and calcifications of the aortic and mitral valves consists of two steps. Each step uses a CNN to classify voxels in the image. The first CNN (\CNN{1}) has a large receptive field to be able to detect calcium based on the anatomical context and to label calcium voxels according to their anatomical location. The second CNN (\CNN{2}) has a smaller receptive field and discards false positives based on local image information. Only voxels that \CNN{1} considers calcium are classified by \CNN{2} as either true-positive or false-positive (\Cref{fig:Flowchart}).

\subsection{First stage network (\CNN{1})}

\begin{figure}
	\includegraphics[width=0.5\textwidth,trim={0 0 2cm 0},clip]{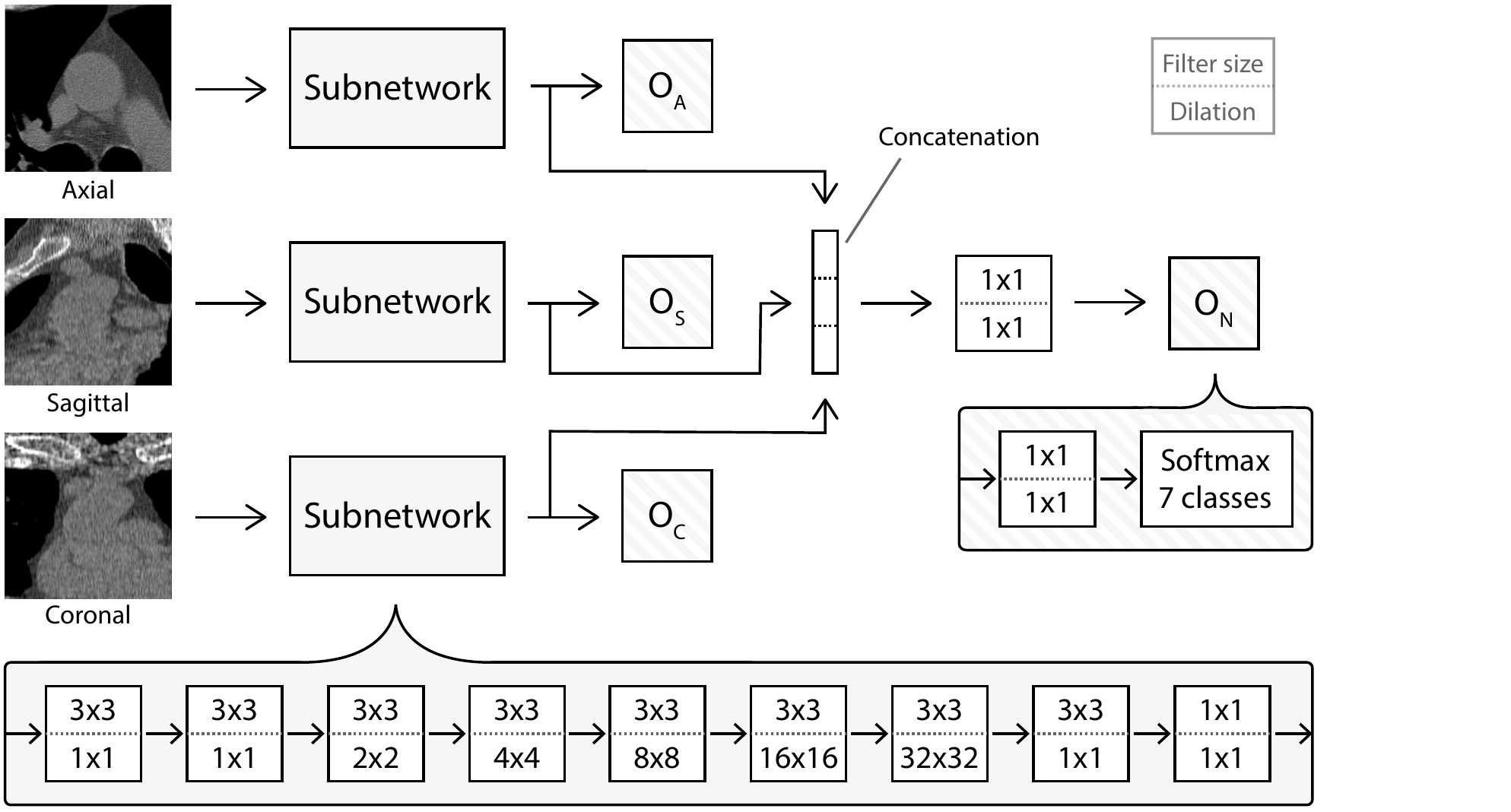}  
	\caption{Architecture of \CNN{1}. The three orthogonal patches are analyzed by subnetworks with identical structure (bottom). The seven output classes are LAD, LCX, RCA, TAC, aortic valve, mitral valve and background. Convolutional layers are shown as boxes with filter size (top) and dilation factor (bottom). All convolutional layers consist of 32 filters, only the layer before $O_N$ consists of 128 filters.}
	\label{fig:CNN1}
\end{figure}

\CNN{1} classifies all voxels in the image that exceed the standard calcium threshold of \SI{130}{HU}. The number of voxels that need to be classified in each scan is therefore high. Classification voxel-by-voxel with a sliding window approach would be highly inefficient as many identical convolutional operations would be repeated unnecessarily. \CNN{1} is therefore constructed as a purely convolutional network \cite{Long2015,Springenberg2015}, i.e., with all layers implemented as convolutions, which allows arbitrary-sized inputs so that entire slices or volumes can be classified at once (\Cref{fig:CNN1}). \CNN{1} classifies voxels as either LAD (including the left main coronary artery), LCX, RCA, TAC, aortic valve calcification, mitral valve calcification or background.

Previous publications showed that spatial information is particularly important for calcium detection. To allow \CNN{1} to infer spatial information from the image area covered by its receptive field, its receptive field needs to be relatively large. However, CNNs with large receptive fields, such as very deep networks or networks with large convolution kernels, often suffer from overfitting due to large numbers of trainable parameters.
To allow for a large receptive field while keeping the number of trainable parameters low, we rely on dilated convolutions, which are based on convolution kernels with spacing between their elements. By stacking convolutions with exponentially growing dilation, the receptive field of the network grows exponentially while the number of trainable parameters only grows linearly\cite{Yu2016}. At the same time, the network still includes all information within its receptive field in the analysis.

The architecture of \CNN{1} is similar to the network proposed by Yu et al.\cite{Yu2016}, but extends it from 2D inputs to three orthogonal 2D inputs (often referred to as 2.5D) \cite{Prasoon2013}. We chose 2.5D inputs over 3D inputs because of the previously reported superior performance for calcium scoring \cite{Wolterink2016}. The receptive field of \CNN{1} is \num{131 x 131} pixels, which corresponds to roughly a quarter of an axial slice.

The input of \CNN{1} is a set of three orthogonal patches, which always intersect in a single voxel regardless of the patch size. This contradicts the idea of purely convolutional networks as larger inputs do then not lead to larger outputs. However, the processing of 2.5D input patches can be divided into 2D subtasks by processing the axial, sagittal and coronal inputs independently\cite{Wolterink2016}. Each 2D input is first separately processed with the subnetwork for the respective orientation, allowing us to obtain a feature representation per patch in an efficient manner. The remaining layers of the network are applied to the concatenated feature vectors from all three orientations to obtain posterior probabilities for each voxel.

Inspired by the concept of deep supervision\cite{Lee2015}, each subnetwork has an auxiliary softmax output layer ($O_A$, $O_S$ and $O_C$ in \Cref{fig:CNN1}). These are used during training to enable learning from input patches larger than the receptive field, i.e., learning from multiple labeled pixels per patch. Using these auxiliary output layers, auxiliary loss terms $L_A$, $L_S$ and $L_C$ are defined. A loss term $L_N$ is defined with the output of the entire network for only the voxel in the intersection of the three orthogonal input patches. These loss terms are combined into an overall loss term $L$ using a weight factor $\gamma$: $L = L_N + \gamma \left( L_A + L_S + L_C \right)$.

After training, the auxiliary output layers are used together with the output layer of the entire network for classification. The corresponding posterior probabilities $\vec{p}_A$, $\vec{p}_S$ and $\vec{p}_C$ from the auxiliary output layers are combined with the posterior probabilities $\vec{p}_N$ by computing the weighted average of the probabilities with weights $\omega_A$, $\omega_S$, $\omega_C$, and $\omega_N$.

\subsection{Second stage network (\CNN{2})}

\CNN{1} detects potential calcifications based on appearance and spatial context, and furthermore determines whether the calcification is located in the coronary arteries (LAD including left main, LCX or RCA), the aorta, or the aortic or mitral valve. However, metal artifacts, image noise or other high intensity structures, such as parts of the spine in direct proximity to the aorta, can result in false positive voxel detections.

\begin{figure}
	\includegraphics[width=0.5\textwidth,trim={0 0 2cm 0},clip]{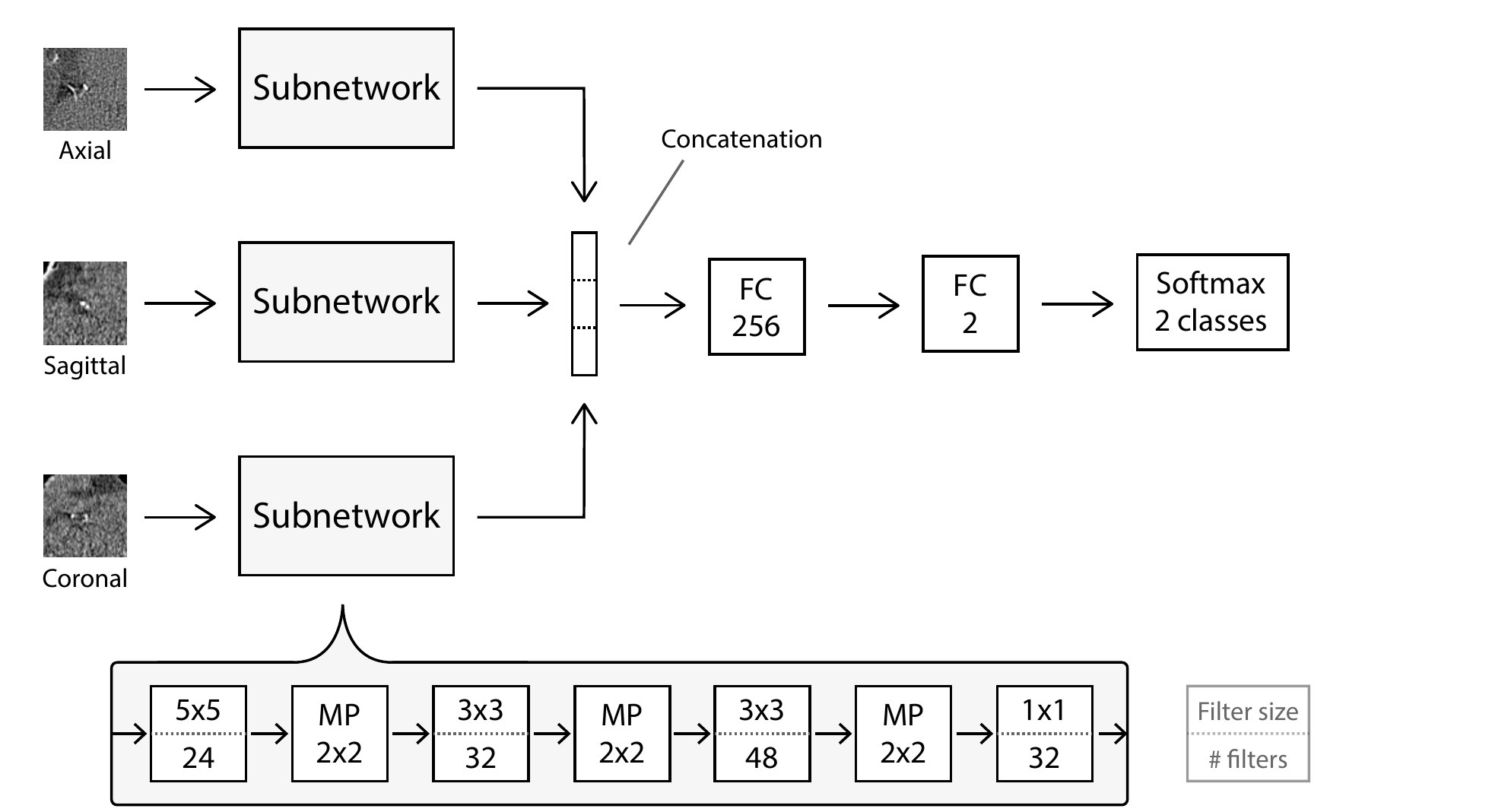}
	\caption{Architecture of \CNN{2}. The three orthogonal patches are analyzed by subnetworks with identical structure (bottom). The two output classes are calcium and background. Convolutional layers are shown as boxes with filter size (top) and number of filters (bottom). \emph{MP} denotes max-pooling layers with the specified pooling region. \emph{FC} denotes fully-connected (dense) layers with the specified number of units.}
	\label{fig:CNN2}
\end{figure}

\CNN{2} refines the output of \CNN{1} by distinguishing between true calcifications and false positive voxels with similar appearance and location. In contrast to \CNN{1}, \CNN{2} does therefore not need to focus on the spatial context but can focus on local information and finer details. \CNN{2} does not use dilated convolutions, but instead non-dilated convolutions with max-pooling between convolutions. \CNN{2} is not purely convolutional like \CNN{1} as it only needs to analyze a limited number of voxels. \CNN{2} analyzes 2.5D inputs and has a receptive field of \num{65 x 65} pixels. Opposed to the multi-class output of \CNN{1}, the output of \CNN{2} is binary as its purpose is false positive reduction and not categorization of the detected calcifications. The architecture of \CNN{2} (\Cref{fig:CNN2}) is motivated by our preliminary work on coronary calcium scoring\cite{Lessmann2016}, in which a similar network achieved good performance when the problem was restricted to a region of interest. In this work, \CNN{1} detects and labels candidate voxels, i.e., objects of interest, instead of a region of interest.

\section{Evaluation}

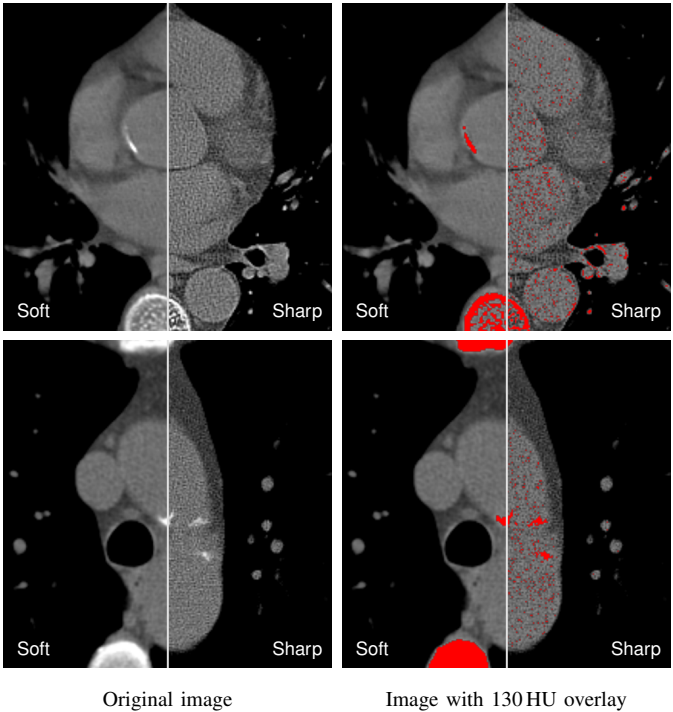
\begin{figure}
	\begin{overpic}[width=0.24\textwidth]{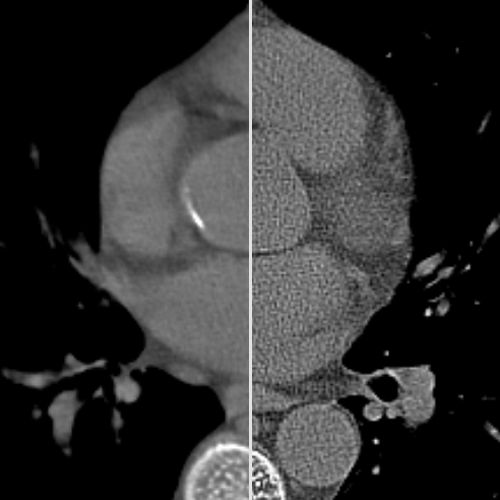}
		\put(4,4){\onimagecaption{Soft}}
		\put(82,4){\onimagecaption{Sharp}}
	\end{overpic}%
	\hfill%
	\begin{overpic}[width=0.24\textwidth]{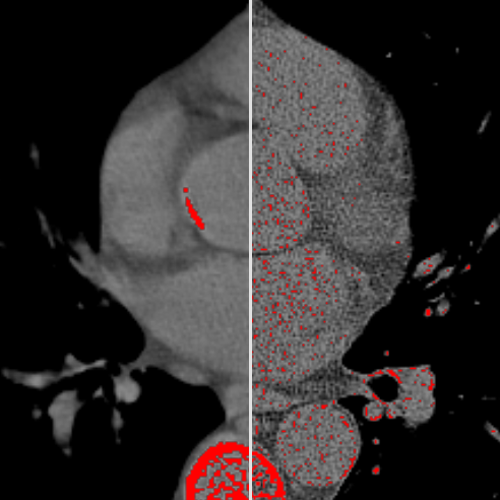}
		\put(4,4){\onimagecaption{Soft}}
		\put(82,4){\onimagecaption{Sharp}}
	\end{overpic}%
	\\[0.5em]
	\begin{overpic}[width=0.24\textwidth]{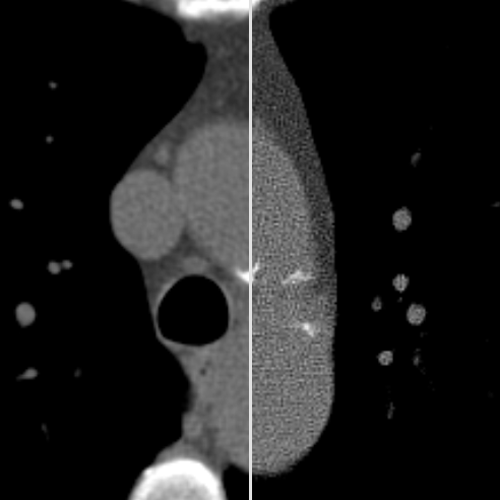}
		\put(4,4){\onimagecaption{Soft}}
		\put(82,4){\onimagecaption{Sharp}}
	\end{overpic}%
	\hfill%
	\begin{overpic}[width=0.24\textwidth]{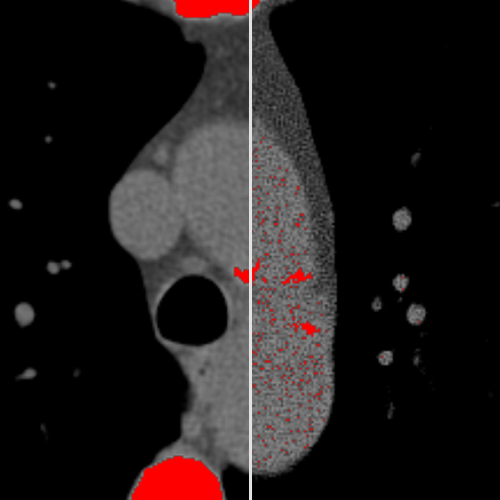}
		\put(4,4){\onimagecaption{Soft}}
		\put(82,4){\onimagecaption{Sharp}}
	\end{overpic}%
	\\[0.25em]
	\makebox[0.24\textwidth]{\footnotesize Original image}%
	\hfill%
	\makebox[0.24\textwidth]{\footnotesize Image with \SI{130}{HU} overlay}%
	\caption{Comparison of low-dose chest CT scans reconstructed with soft filter kernel (left half of each image) and sharp filter kernel (right half of each image). The figures on the right show the same scans as the figures on the left, but with voxels $\ge$\SI{130}{HU} highlighted in red, indicating all voxels above the standard calcium threshold.}
	\label{fig:SoftSharpExamples}
\end{figure}

Calcium scoring is normally performed in images with soft tissue reconstruction. However, lung cancer screening data also includes images reconstructed with sharper filter kernels, in which edges but also noise appear more prominent (\Cref{fig:SoftSharpExamples}). To evaluate whether our method needs to be trained with images that are reconstructed with parameters recommended for calcium scoring (soft tissue filter kernels) or whether it can be trained with all images acquired in the screening (see \Cref{tbl:ScannerModels}), two pairs of \CNN{1} and \CNN{2} were trained: one pair using soft reconstructions only and the other pair using both soft and sharp reconstructions. Furthermore, to evaluate whether calcium detection can be performed in images reconstructed using soft and sharp filter kernels, and to evaluate which of the two training settings leads to best performance, we evaluated our method separately on soft and sharp reconstructions.

To evaluate the performance of the method, calcifications were quantified per subject and per label using volume and Agatston scores. Agatston scores were normalized to account for overlapping slices \cite{Ohnesorge2002}. The agreement between automatically and manually determined calcium volumes was assessed using sensitivity, average false positive volume per scan and the F$_1$ score. Interobserver agreement was assessed using the same metrics by comparing the annotations of the second and third observers to the radiologist as reference.

Additionally, we evaluated cardiovascular risk classification. Each subject was assigned one of four risk categories (\rom{1}--\rom{4}: 0--10, 11--100, 101--1000, >1000) based on their total CAC Agatston score. Reliability of the risk category assignment was assessed using the linearly weighted $\kappa$ coefficient.

\begin{table*}
	\centering
	\caption{Manual and automatic CAC scoring performance on 310~soft and 196~sharp reconstructions. \emph{Reference standard} refers to the amount of calcium manually identified by the observers. The performance of a \emph{second and third observer} on a subset of 100 scans is additionally reported. The performance of the automatic method is reported when the networks were \emph{trained on soft reconstructions} and when they were \emph{trained on soft and sharp reconstructions}.}
	\label{tbl:Results_CAC}
	\begin{tabular}{lcrrrrcrrrr}
		\toprule
		& & \multicolumn{4}{c}{\emph{Soft reconstructions}} & & \multicolumn{4}{c}{\emph{Sharp reconstructions}} \\
		\addlinespace[0.15em]
		& & CAC & LAD & LCX & RCA & & CAC & LAD & LCX & RCA \\
		\midruleheading{11}{Reference standard}
		Scans with any calcification (\si{\percent}) & & \num{74.5} & \num{67.7} & \num{41.6} & \num{48.1} & & \num{76.5} & \num{71.9} & \num{36.7} & \num{49.5}\\
		Calcium volume / scan (\si{\milli\meter\cubed}) & & \num{303.4} & \num{150.2} & \num{50.4} & \num{102.9} & & \num{377.65} & \num{171.3} & \num{51.1} & \num{155.3}\\
		\midruleheading{11}{Second observer\textsuperscript{\dag}}
		Sensitivity (\si{\percent}) & & \num{94.8} & \num{93.5} & \num{83.6} & \num{95.8} & & \num{86.8} & \num{88.5} & \num{79.4} & \num{84.1}\\
		False positive volume / scan (\si{\milli\meter\cubed}) & & \num{20.9} & \num{9.9} & \num{9.2} & \num{12.3} & & \num{23.3} & \num{15.0} & \num{9.2} & \num{9.7}\\
		F$_1$ score calcium volume & & \num{0.95} & \num{0.94} & \num{0.87} & \num{0.96} & & \num{0.85} & \num{0.86} & \num{0.75} & \num{0.83}\\
		\midruleheading{11}{Third observer\textsuperscript{\dag}}
		Sensitivity (\si{\percent}) & & \num{97.6} & \num{98.9} & \num{87.0} & \num{96.7} & & \num{91.0} & \num{91.2} & \num{86.0} & \num{87.8}\\
		False positive volume / scan (\si{\milli\meter\cubed}) & & \num{19.7} & \num{12.3} & \num{8.1} & \num{10.2} & & \num{13.5} & \num{7.3} & \num{11.1} & \num{8.5}\\
		F$_1$ score calcium volume & & \num{0.97} & \num{0.96} & \num{0.89} & \num{0.97} & & \num{0.94} & \num{0.94} & \num{0.86} & \num{0.92}\\
		\midruleheading{11}{Trained on soft reconstructions only}
		Sensitivity (\si{\percent}) & & \num{91.3} & \num{92.5} & \num{71.8} & \num{91.5} & & \num{53.7} & \num{54.0} & \num{44.9} & \num{54.1}\\
		False positive volume / scan (\si{\milli\meter\cubed}) & & \num{35.5} & \num{18.3} & \num{13.6} & \num{11.2} & & \num{19.4} & \num{10.6} & \num{4.7} & \num{7.4}\\
		F$_1$ score calcium volume & & \num{0.90} & \num{0.90} & \num{0.72} & \num{0.90} & & \num{0.68} & \num{0.67} & \num{0.58} & \num{0.68}\\
		\midruleheading{11}{Trained on soft and sharp reconstructions}
		Sensitivity (\si{\percent}) & & \num{91.2} & \num{93.7} & \num{65.8} & \num{90.4} & & \num{84.4} & \num{86.9} & \num{61.6} & \num{83.0}\\
		False positive volume / scan (\si{\milli\meter\cubed}) & & \num{40.7} & \num{26.6} & \num{12.0} & \num{11.8} & & \num{62.8} & \num{37.0} & \num{9.9} & \num{25.2}\\
		F$_1$ score calcium volume & & \num{0.89} & \num{0.89} & \num{0.69} & \num{0.90} & & \num{0.84} & \num{0.83} & \num{0.68} & \num{0.83}\\
		\bottomrule
		\addlinespace[0.35em]
		\multicolumn{11}{r}{\textsuperscript{\dag}subset of 100 scans}
	\end{tabular}
\end{table*}

\bgroup
\def\arraystretch{1.3}%
\begin{table*}
	\centering
	\caption{Agreement in risk categorization of subjects based on their total CAC Agatston score (\rom{1}: 0--10, \rom{2}: 11--100, \rom{3}: 101--1000, \rom{4}: >1000) between manual reference standard and automatically determined scores. The agreement is reported separately for soft and sharp reconstructions. The table on the left side specifies the agreement when the automatic method was \emph{trained on soft reconstructions only} and the table on the right side when the automatic method was \emph{trained on soft and sharp reconstructions}.}
	\label{tbl:RiskCategoryAgreement}
	\begin{tabular}{lc*{4}{c}c*{4}{c}}\hline
		\addlinespace[0.15em]
		& & \multicolumn{4}{c}{\emph{Soft reconstructions}} & & \multicolumn{4}{c}{\emph{Sharp reconstructions}}\\
		\addlinespace[0.1em]
		& & \multicolumn{4}{c}{Automatic} & & \multicolumn{4}{c}{Automatic}\\
		Reference & & \rom{1} & \rom{2} & \rom{3} & \rom{4} & & \rom{1} & \rom{2} & \rom{3} & \rom{4}\\\hline
		\rom{1} & & \textbf{\num{90}} & \num{17} & \num{1} & \num{0} & & \textbf{\num{62}} & \num{2} & \num{1} & \num{0}\\
		\rom{2} & & \num{3} & \textbf{\num{59}} & \num{4} & \num{0} & & \num{12} & \textbf{\num{18}} & \num{0} & \num{0}\\
		\rom{3} & & \num{0} & \num{2} & \textbf{\num{99}} & \num{2} & & \num{9} & \num{12} & \textbf{\num{51}} & \num{1}\\
		\rom{4} & & \num{0} & \num{0} & \num{1} & \textbf{\num{32}} & & \num{3} & \num{4} & \num{6} & \textbf{\num{15}}\\
		\hline
		& & \multicolumn{4}{c}{$\kappa = 0.91$} & & \multicolumn{4}{c}{$\kappa = 0.70$}\\
		\addlinespace[0.25em]
		\multicolumn{11}{c}{\emph{Trained on soft reconstructions only}}
	\end{tabular}
	\hspace{3em}
	\begin{tabular}{lc*{4}{c}c*{4}{c}}\hline
		\addlinespace[0.15em]
		& & \multicolumn{4}{c}{\emph{Soft reconstructions}} & & \multicolumn{4}{c}{\emph{Sharp reconstructions}}\\
		\addlinespace[0.1em]
		& & \multicolumn{4}{c}{Automatic} & & \multicolumn{4}{c}{Automatic}\\
		Reference & & \rom{1} & \rom{2} & \rom{3} & \rom{4} & & \rom{1} & \rom{2} & \rom{3} & \rom{4}\\\hline
		\rom{1} & & \textbf{\num{92}} & \num{12} & \num{4} & \num{0} & & \textbf{\num{55}} & \num{10} & \num{0} & \num{0}\\
		\rom{2} & & \num{3} & \textbf{\num{58}} & \num{5} & \num{0} & & \num{0} & \textbf{\num{25}} & \num{5} & \num{0}\\
		\rom{3} & & \num{0} & \num{2} & \textbf{\num{99}} & \num{2} & & \num{1} & \num{1} & \textbf{\num{68}} & \num{3}\\
		\rom{4} & & \num{0} & \num{0} & \num{1} & \textbf{\num{32}} & & \num{0} & \num{0} & \num{1} & \textbf{\num{27}}\\
		\hline
		& & \multicolumn{4}{c}{$\kappa = 0.91$} & & \multicolumn{4}{c}{$\kappa = 0.90$}\\
		\addlinespace[0.25em]
		\multicolumn{11}{c}{\emph{Trained on soft and sharp reconstructions}}
	\end{tabular}
\end{table*}
\egroup

\section{Experiments and Results} \label{sec:Results}

The image quality of 57 scans (\SI{3}{\percent}) was considered inadequate for manual annotation due to severe metal artifacts or excessive image noise. The remaining \num{1687} scans with manual reference standard were divided into subsets for training (\SI{60}{\percent}\,=\,1012 scans), validation (\SI{10}{\percent}\,=\,169 scans) and testing (\SI{30}{\percent}\,=\,506 scans). These were scans of \num{1459} different participants as sometimes multiple reconstructions of the same baseline scan were included. The division into subsets was random, but all scans of the same participant were assigned to the same subset. Among the \num{1687} scans, \SI{58}{\percent} were reconstructed with soft filters and \SI{42}{\percent} with sharp filters. Among the \num{100} scans with multiple annotations, \SI{52}{\percent} were reconstructed with soft filters and \SI{48}{\percent} with sharp filters.

The areas covered by the receptive fields of both networks are not necessarily comparable across scans due to different resolutions. We therefore resampled all scans in-plane with bilinear interpolation to \SI{0.66 x 0.66}{\milli\meter}, which is the average resolution across the dataset. Resampling was only performed in-plane because the slice spacing was already standardized to \SI{1.5}{\milli\meter} before manual annotation (see \Cref{sec:Data}). Both networks \CNN{1} and \CNN{2} thus analyzed images at a standardized resolution. The predicted label maps were finally resampled to the original resolution using nearest neighbor interpolation.

The two networks \CNN{1} and \CNN{2} were trained sequentially. \CNN{1} was trained with high density voxels ($\ge$\SI{130}{HU}) in the training scans. \CNN{2} was trained with high density voxels classified as any type of calcification by \CNN{1}. The validation set was used to ensure there was no substantial overfitting and to determine convergence of the networks. We trained both networks on balanced minibatches, which consisted half of randomly selected calcium voxels of any class and half of randomly selected background voxels.
Both networks used exponential linear units\cite{Clevert2016} as activation function. Adam\cite{Kingma2014} was used as optimizer with a learning rate of \num{5e-4} and the categorical cross-entropy as loss function. \CNN{1} was trained on patches of \num{155x155} pixels, i.e., larger than its receptive field, with $\gamma = 0.05$ and \CNN{2} on patches of \num{65 x 65} pixels. During training of both networks, Dropout\cite{Srivastava2014} (\CNN{1}: \SI{35}{\percent}, \CNN{2}: \SI{50}{\percent}) and L2 weight decay (\CNN{1}: \num{5e-5}, \CNN{2}: \num{1e-5}) were used for regularization. Since \CNN{2} has many more trainable parameters, we added batch normalization\cite{Ioffe2015} between all layers to provide additional regularization. For both networks, the output class was determined as the class with highest activation. \CNN{1} used $\omega_N = \frac{1}{2}$ and $\omega_{A,S,C} = \frac{1}{6}$ to weight the output layers, which corresponds to averaging between the normal and the auxiliary outputs.

The networks were implemented using the Theano\cite{Theano} and Lasagne\cite{Lasagne} frameworks and trained on NVIDIA Titan X GPUs. The total computation time is 5--7 minutes, depending on the size of the image volume and the number of candidate objects in the image. With our non-optimized implementation, \CNN{1} needs on average $5.5 \pm 1$ minutes to scan an entire image, and \CNN{2} needs on average $8.6 \pm 8.1$ seconds to classify the detected candidate voxels.

\subsection{Detection of CAC}

The performance of automatic CAC detection was evaluated based on scores per artery and per subject.
Per artery and per subject sensitivities, average false positive volumes and F$_1$ scores for CAC detection are listed in \Cref{tbl:Results_CAC}. Examples of detected calcifications are shown in \Cref{fig:Examples}.

In images reconstructed with soft filter kernels, the automatic method detected more than \SI{90}{\percent} of the calcium in LAD and RCA, and \SI{72}{\percent} in LCX. Training on soft and sharp reconstructions compared to only soft reconstructions led to similar performance for CAC with F$_1$ scores of \num{0.89} and \num{0.90}. In images reconstructed with sharp filter kernels, the F$_1$ score for CAC increased from \num{0.68} to \num{0.84} when sharp reconstructions were added to the training data.

Risk categories derived from per-subject CAC scores agreed with the manual reference annotation in images reconstructed with soft filter kernels in \SI{90}{\percent} of the subjects when trained only on soft reconstructions and in \SI{91}{\percent} when sharp reconstructions were added to the training data. In images reconstructed with sharp filter kernels, agreement increased from \SI{75}{\percent} to \SI{89}{\percent} when sharp reconstructions were added to the training data. Confusion matrices for the risk category assignment are shown in Table~\rom{3}.

Interobserver agreement was high for per-subject CAC in soft reconstructions with F$_1$ scores of \num{0.95} and \num{0.97} for the second and third observer, respectively. Interobserver agreement was overall slightly lower in sharp reconstructions compared to soft reconstructions. For risk category assignment, $\kappa$ was \num{0.98} and \num{1.00} for the second and third observer, respectively, in soft reconstructions and \num{0.92} and \num{0.99} in sharp reconstructions.

\subsection{Detection of TAC}

\begin{table}
	\centering
	\caption{Manual and automatic TAC scoring performance on 310~soft and 196~sharp reconstructions. \emph{Reference standard} refers to the amount of calcium manually identified by the observers. The performance of a \emph{second and third observer} on a subset of 100 scans is additionally reported. The performance of the automatic method is reported when the networks were \emph{trained on soft reconstructions} and when they were \emph{trained on soft and sharp reconstructions}.}
	\label{tbl:Results_TAC}
	\begin{tabular}{lrr}
		\toprule
		Reconstruction filter & \emph{Soft} & \emph{Sharp}\\
		\midruleheading{3}{Reference standard}
		Scans with any calcium (\si{\percent}) & \num{89.4} & \num{88.3}\\
		Calcium volume / scan (\si{\milli\meter\cubed}) & \num{1875.0} & \num{1476.4}\\
		\midruleheading{3}{Second observer\textsuperscript{\dag}}
		Sensitivity (\si{\percent}) & \num{97.6} & \num{90.2}\\
		False positive volume / scan (\si{\milli\meter\cubed}) & \num{300.6} & \num{367.7}\\
		F$_1$ score calcium volume & \num{0.93} & \num{0.79}\\
		\midruleheading{3}{Third observer\textsuperscript{\dag}}
		Sensitivity (\si{\percent}) & \num{93.7} & \num{86.5}\\
		False positive volume / scan (\si{\milli\meter\cubed}) & \num{366.2} & \num{537.4}\\
		F$_1$ score calcium volume & \num{0.90} & \num{0.75}\\
		\midruleheading{3}{Trained on soft reconstructions only}
		Sensitivity (\si{\percent}) & \num{89.8} & \num{60.1}\\
		False positive volume / scan (\si{\milli\meter\cubed}) & \num{213.3} & \num{152.7}\\
		F$_1$ score calcium volume & \num{0.89} & \num{0.71}\\
		\midruleheading{3}{Trained on soft and sharp reconstructions}
		Sensitivity (\si{\percent}) & \num{92.0} & \num{85.1}\\
		False positive volume / scan (\si{\milli\meter\cubed}) & \num{258.9} & \num{379.3}\\
		F$_1$ score calcium volume & \num{0.89} & \num{0.81}\\
		\bottomrule
		\addlinespace[0.35em]
		\multicolumn{3}{r}{\textsuperscript{\dag}subset of 100 scans}
	\end{tabular}
\end{table}

The observers identified TAC in the majority of scans, in \num{450} of \num{506} (\SI{89}{\percent}). The detection performance of the automatic method in terms of sensitivity, average false positive volume and F$_1$ score is listed in \Cref{tbl:Results_TAC}.
In images reconstructed with soft filter kernels, the automatic method achieved an F$_1$ score of \num{0.89} regardless whether only soft reconstructions or soft and sharp reconstructions were used for training. In images reconstructed with sharp filter kernels, the F$_1$ score increased from \num{0.71} to \num{0.81} when sharp reconstructions were added to the training data. Interobserver agreement was high for TAC and, overall, the observers had higher sensitivity than the automatic method, but also higher average false positive volume.

\subsection{Detection of cardiac valve calcifications}

\begin{table*}
	\centering
	\caption{Performance of manual and automatic scoring of aortic and mitral valve calcifications, reported separately for 310~soft and 196~sharp reconstructions. \emph{Reference standard} refers to the amount of calcium manually identified by the observers. The performance of a \emph{second and third observer} on a subset of 100 scans is additionally reported. The performance of the automatic method is reported when the networks were \emph{trained on soft reconstructions} and when they were \emph{trained on soft and sharp reconstructions}.}
	\label{tbl:Results_Valves}
	\begin{tabular}{lrrcrr}
		\toprule
		& \multicolumn{2}{c}{\emph{Soft reconstructions}} & & \multicolumn{2}{c}{\emph{Sharp reconstructions}}\\
		\addlinespace[0.15em]
		& Aortic valve & Mitral valve & & Aortic valve & Mitral valve\\
		\midruleheading{6}{Reference standard}
		Scans with any calcium (\si{\percent}) & \num{20.6} & \num{12.6} & & \num{14.3} & \num{9.2}\\
		Calcium volume / scan (\si{\milli\meter\cubed}) & \num{44.9} & \num{31.1} & & \num{62.1} & \num{38.8}\\
		\midruleheading{6}{Second observer\textsuperscript{\dag}}
		Sensitivity (\si{\percent}) & \num{86.8} & \num{83.2} & & \num{21.9} & \num{9.0}\\
		False positive volume / scan (\si{\milli\meter\cubed}) & \num{36.5} & \num{8.7} & & \num{4.1} & \num{1.7}\\
		F$_1$ score calcium volume & \num{0.74} & \num{0.88} & & \num{0.18} & \num{0.09}\\
		\midruleheading{6}{Third observer\textsuperscript{\dag}}
		Sensitivity (\si{\percent}) & \num{97.1} & \num{88.0} & & \num{87.3} & \num{9.8}\\
		False positive volume / scan (\si{\milli\meter\cubed}) & \num{64.1} & \num{19.1} & & \num{15.1} & \num{1.6}\\
		F$_1$ score calcium volume & \num{0.69} & \num{0.88} & & \num{0.51} & \num{0.17}\\
		\midruleheading{6}{Trained on soft reconstructions only}
		Sensitivity (\si{\percent}) & \num{55.3} & \num{64.2} & & \num{44.8} & \num{39.4}\\
		False positive volume / scan (\si{\milli\meter\cubed}) & \num{5.4} & \num{18.2} & & \num{3.6} & \num{4.3}\\
		F$_1$ score calcium volume & \num{0.66} & \num{0.58} & & \num{0.59} & \num{0.52}\\
		\midruleheading{6}{Trained on soft and sharp reconstructions}
		Sensitivity (\si{\percent}) & \num{58.7} & \num{67.1} & & \num{57.1} & \num{60.8}\\
		False positive volume / scan (\si{\milli\meter\cubed}) & \num{7.4} & \num{24.5} & & \num{12.9} & \num{9.5}\\
		F$_1$ score calcium volume & \num{0.67} & \num{0.55} & & \num{0.64} & \num{0.66}\\
		\bottomrule
		\addlinespace[0.35em]
		\multicolumn{6}{r}{\textsuperscript{\dag}subset of 100 scans}
	\end{tabular}
\end{table*}

Cardiac valve calcifications were infrequently identified by the observers, aortic valve calcifications in 92 of 506 scans (\SI{18.2}{\percent}) and mitral valve calcifications in 58 of 506 scans (\SI{11.5}{\percent}). The detection performance of the automatic method in terms of sensitivity, average false positive volume and F$_1$ score is listed in \Cref{tbl:Results_Valves}.

The automatic method achieved lower performance for detection of cardiac valve calcifications compared to detection of CAC and TAC. However, similar was that adding sharp reconstructions to the training data improved F$_1$ scores in sharp reconstructions, from \num{0.59} to \num{0.64} for aortic valve calcifications and from \num{0.52} to \num{0.66} for mitral valve calcifications.
Interobserver agreement in terms of F$_1$ score was particularly low in images reconstructed with sharp filter kernels low with F$_1$ scores of \num{0.18}\,/\,\num{0.51} and \num{0.09}\,/\,\num{0.17} for aortic and mitral valve calcifications, respectively.

\subsection{Single vs.\ two-stage performance}

To evaluate the contribution of \CNN{2} to the overall performance, we compared the performance of \CNN{1} followed by \CNN{2} against \CNN{1} alone. Trained using only images with soft reconstruction kernels, \CNN{1} achieved a sensitivity of \SI{96.6}{\percent} with an average false-positive volume of \SI{5574}{\cubic\milli\meter} on soft reconstructions, and \SI{90.7}{\percent} sensitivity with \SI{18739}{\cubic\milli\meter} average false-positive volume on sharp reconstructions. Reclassification of the positive detections by \CNN{2} reduced the sensitivity by \SI{6.1}{\percent} and \SI{32.0}{\percent}, but at the same time reduced the average false-positive volume per scan by \SI{95.7}{\percent} and \SI{99.1}{\percent} in soft and sharp reconstructions, respectively.

When images with sharp reconstruction kernels were added to the training data, \CNN{1} achieved a sensitivity of \SI{97.3}{\percent} at an average false-positive volume of \SI{6495}{\cubic\milli\meter} in soft reconstructions. In sharp reconstructions, the sensitivity was \SI{92.6}{\percent} at \SI{14802}{\cubic\milli\meter} average false-positive volume. Reclassification by \CNN{2} reduced the sensitivity by \SI{4.8}{\percent} and \SI{7.7}{\percent}, but at the same time reduced the average false-positive volume per scan by \SI{95.4}{\percent} and \SI{97.0}{\percent} in soft and sharp reconstructions, respectively.

\subsection{Effect of receptive field size}

To evaluate the influence of the size of the receptive field of \CNN{1} on the detection performance, we trained networks with various maximum dilation factors. A larger maximal dilation factor results in a larger receptive field and also a deeper network (\Cref{fig:CNN1};\cite{Yu2016}). For receptive field sizes of \num{35x35}, \num{67x67}, \num{131x131} and \num{259x259}, \CNN{1} achieved F$_1$ scores of \num{0.27}, \num{0.29}, \num{0.40} and \num{0.15}, respectively, in images reconstructed with soft reconstruction kernels. In images with sharp reconstruction kernel, the F$_1$ scores were \num{0.18}, \num{0.14}, \num{0.20} and \num{0.07}. The largest network had to be trained on smaller batches of samples due to hardware limitations (32 instead of 64), which made the network more difficult to train.

\subsection{Effect of auxiliary output layers and loss terms}

We additionally evaluated whether the auxiliary softmax output layers of \CNN{1} with the corresponding auxiliary loss terms had a positive effect on training time and performance. We observed that the network learned slower with the auxiliary loss terms. However, the detection performance improved. The performance of \CNN{1} with and without auxiliary output layers and loss terms was compared using training images reconstructed with both soft and sharp reconstruction kernels. In images reconstructed with soft filter kernels, F$_1$ scores for calcium detection (binary, disregarding the label) were \num{0.17} without auxiliary outputs and \num{0.40} with auxiliary outputs. In images reconstructed with sharp filter kernels, the F$_1$ scores were \num{0.07} without auxiliary outputs and \num{0.20} with auxiliary outputs.

\subsection{Comparison with other methods}

The number of other publications on automatic calcium scoring methods in low-dose chest CT is low. No other methods have been published that concurrently detected CAC, TAC and cardiac valve calcifications, and also no methods for detection of cardiac valve calcifications. Hence, the results of the proposed combined system can only be compared to the simpler tasks of detecting either only CAC or only TAC.

For automatic CAC scoring in low-dose chest CT, I\v{s}gum~et~al.\cite{Isgum2012} reported a sensitivity of \SI{79}{\percent} at \SI{4}{\cubic\milli\meter} false positive volume per scan in \num{231} scans. In the same scans, Lessmann~et~al.\cite{Lessmann2016} achieved a sensitivity of \SI{97}{\percent} at an average false positive volume of \SI{10}{\cubic\milli\meter}.
However, the dataset on which these methods were tested was much less diverse than the dataset used in this paper. All scans were acquired in the same hospital with CT scanners from a single vendor and reconstructed using a single soft filter kernel. Furthermore, the average CAC burden per subject was considerably lower (\SI{198}{\cubic\milli\meter} vs.\ \SI{303}{\cubic\milli\meter}). We therefore evaluated the better performing method\cite{Lessmann2016} on the current test data after retraining with the current training data. The method described in \cite{Lessmann2016} classifies individual voxels, but originally the average posterior probabilities across connected voxels was calculated to classify lesions rather than individual voxels. This averaging was now omitted to allow for a comparison on voxel-level. The best performance was achieved on soft reconstructions, using both soft and sharp reconstructions for training: the average F$_1$ score per scan for CAC detection was \num{0.23} (\SI{95}{\percent} CI: \num{0.20}--\num{0.25}), mostly attributed to many false positive detections. In comparison, the proposed method achieved an average F$_1$ score of \num{0.87} (\SI{95}{\percent} CI: \num{0.85}--\num{0.90}) on the same test data using the same training data.
This difference was statistically significant ($p < 0.001$, paired samples t-test).

Other methods for automatic CAC scoring were only evaluated in terms of their correlations with manual scores: Xie~et~al.\cite{Xie2014} performed linear regression with CAC Agatston scores in \num{41} scans and reported $R^2 = 0.91$. González~et~al.\cite{Gonzalez2016} reported a Pearson correlation coefficient of \num{0.86} for CAC Agatston scores in \num{1000} scans.

For automatic TAC scoring in low-dose chest CT, I\v{s}gum~et~al.\cite{Isgum2010} reported a sensitivity of \SI{98}{\percent} at \SI{64}{\cubic\milli\meter} false positive volume per scan in \num{93} scans.
Kurugol~et~al.\cite{Kurugol2012} reported a sensitivity of \SI{94}{\percent} and a positive predictive value of \SI{91}{\percent} for TAC volume in \num{45} scans. In comparison, we reported here a sensitivity of \SI{90}{\percent} at an average false positive volume of \SI{213}{\cubic\milli\meter} and a positive predictive value of \SI{89}{\percent} in \num{310} scans reconstructed with soft filter kernels. Xie~et~al.\cite{Xie2014b} reported only the correlation with manual scores as $R^2 = 0.98$ after performing linear regression for TAC volume in \num{45} scans.


\begin{figure*}
	\centering
	\includegraphics[width=0.24\textwidth]{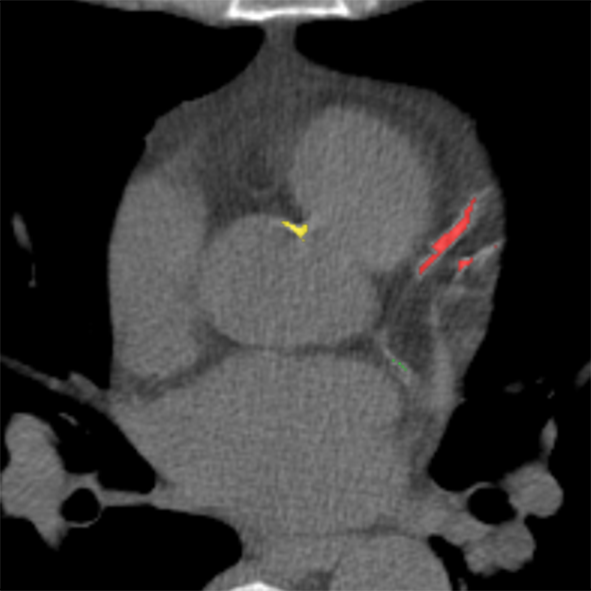}
	\includegraphics[width=0.24\textwidth]{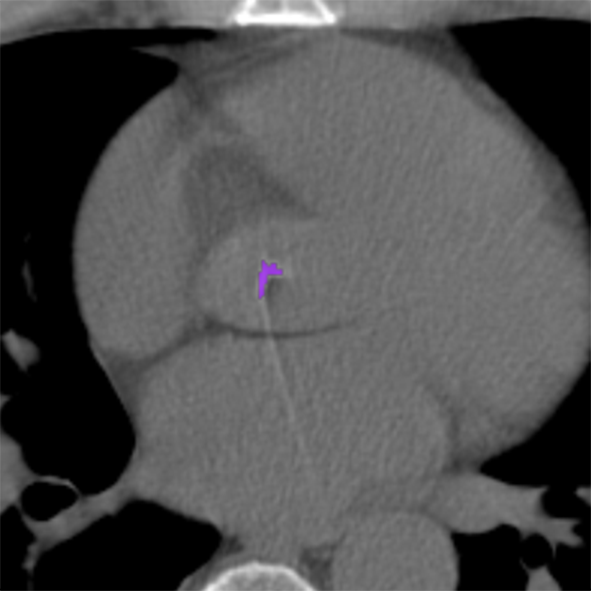}
	\includegraphics[width=0.24\textwidth]{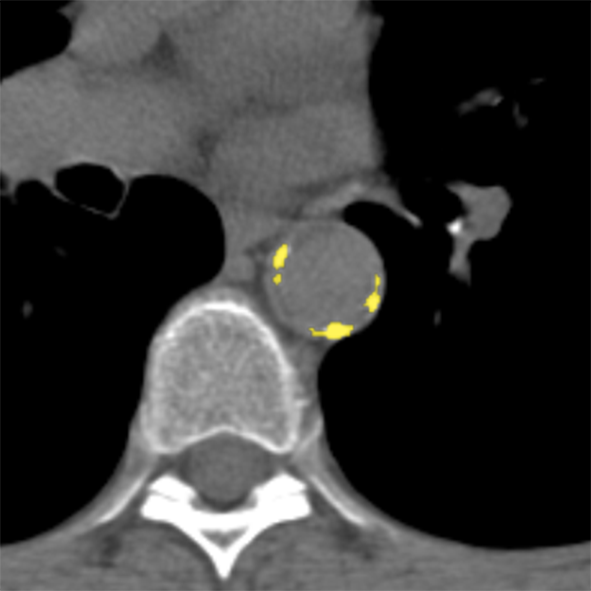}
	\includegraphics[width=0.24\textwidth]{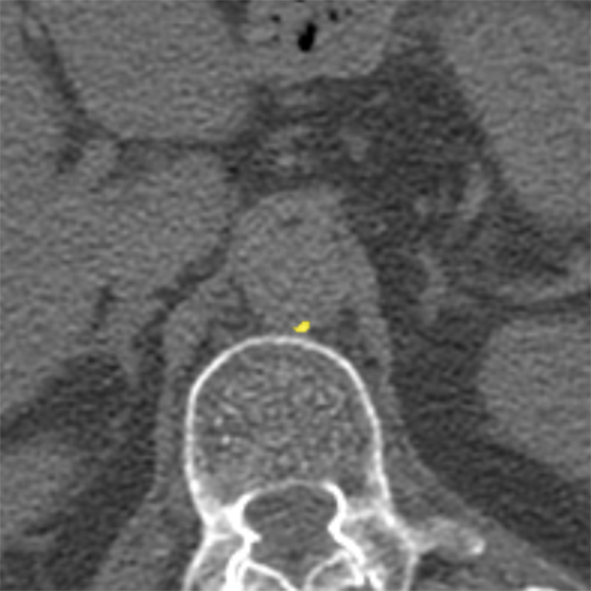}%
	\\
	\quartercaption{(a)}
	\quartercaption{(b)}
	\quartercaption{(c)}
	\quartercaption{(d)}%
	\\[0.5em]
	\includegraphics[width=0.24\textwidth]{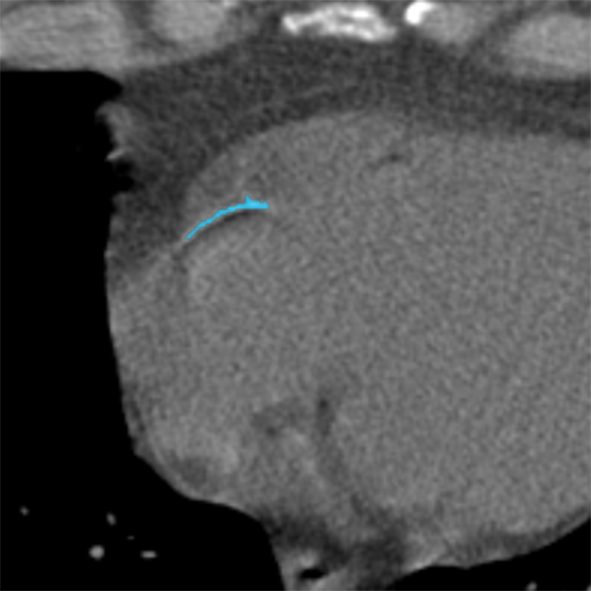}
	\includegraphics[width=0.24\textwidth]{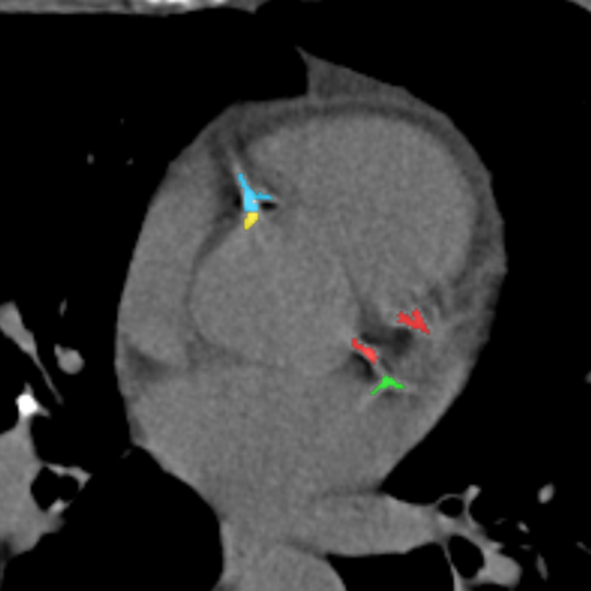}	
	\includegraphics[width=0.24\textwidth]{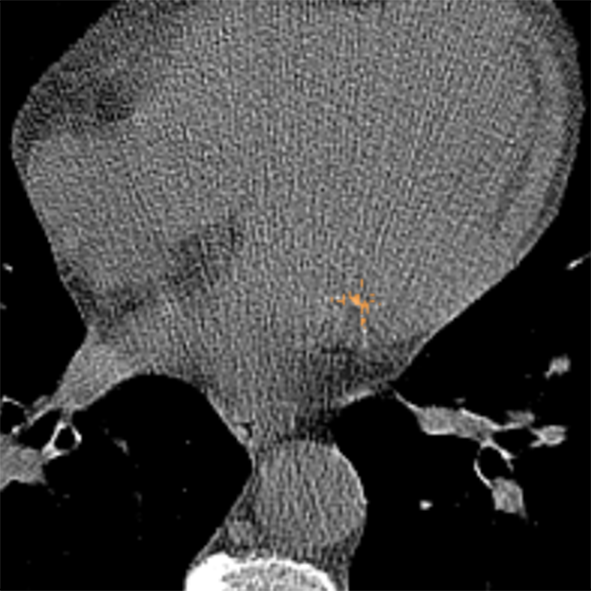}
	\includegraphics[width=0.24\textwidth]{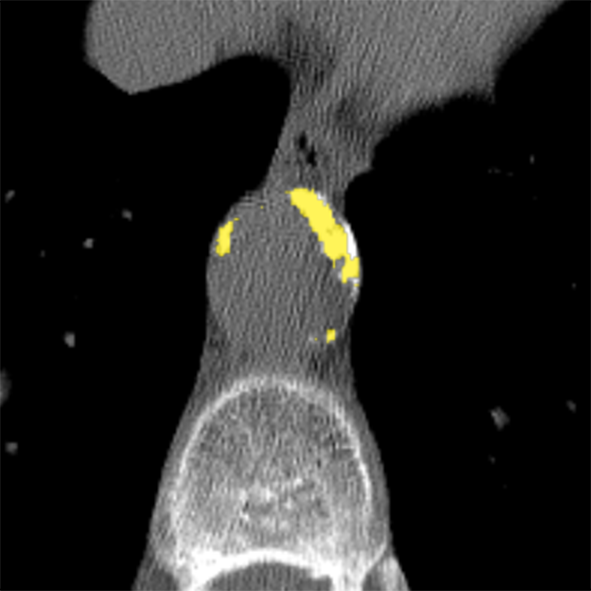}%
	\\
	\quartercaption{(e)}
	\quartercaption{(f)}
	\quartercaption{(g)}
	\quartercaption{(h)}%
	\\[0.5em]	
	\includegraphics[width=0.24\textwidth]{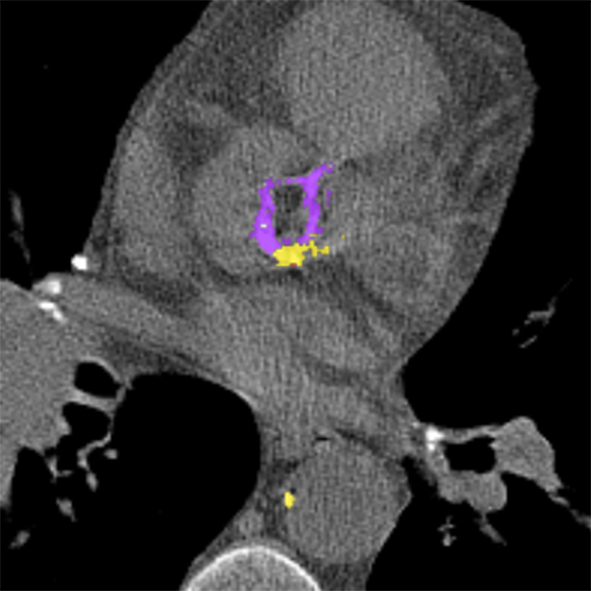}
	\includegraphics[width=0.24\textwidth]{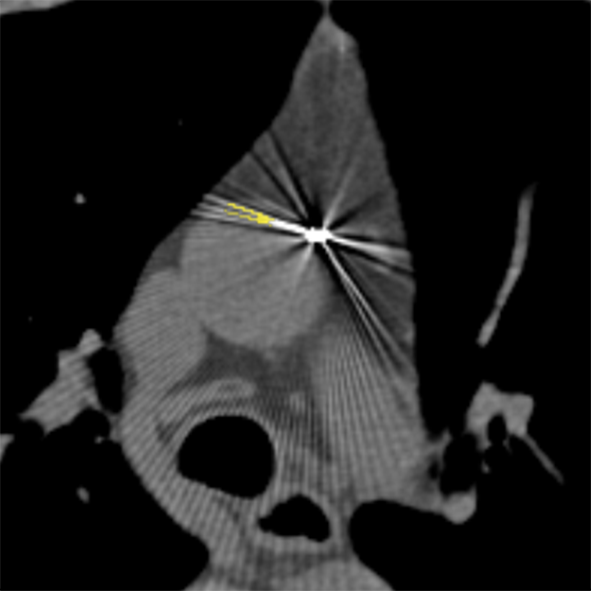}
	\includegraphics[width=0.24\textwidth]{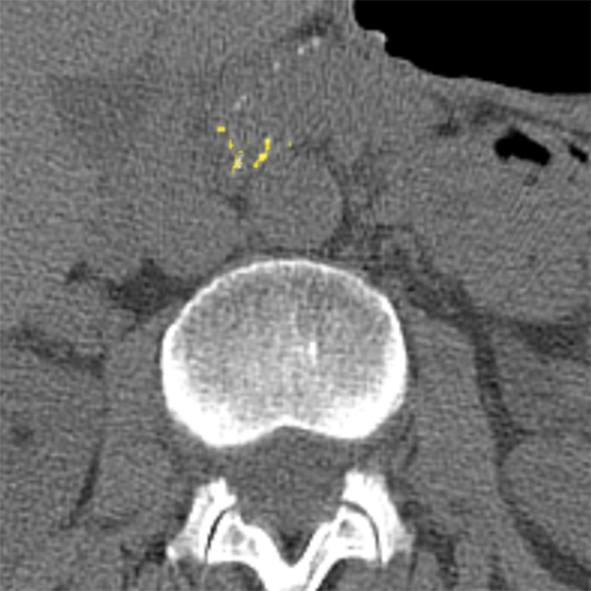}	
	\includegraphics[width=0.24\textwidth]{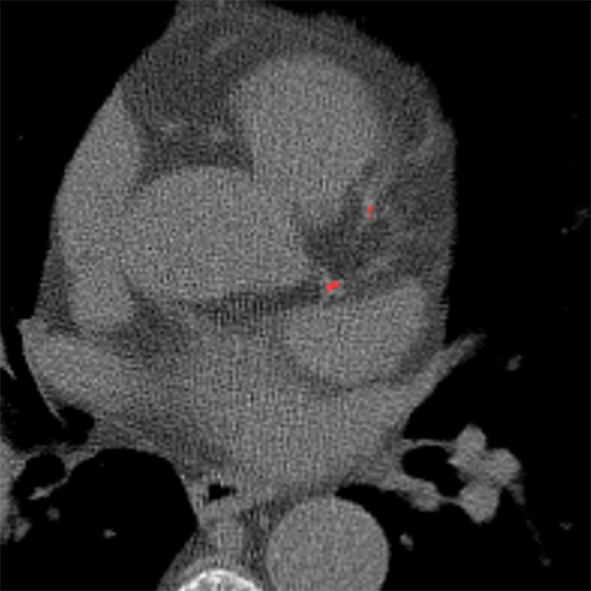}%
	\\
	\quartercaption{(i)}
	\quartercaption{(j)}
	\quartercaption{(k)}
	\quartercaption{(l)}%
	\caption{Example cases overlayed with color coded automatic detection results. The color scheme is as follows: red=LAD including LM, green=LCX, blue=RCA, yellow=TAC, purple=aortic valve calcification, orange=mitral valve calcification. Examples of (a)--(f) correctly detected and labeled calcifications, (g) and (h) correctly labeled calcifications of which a few voxels were missed, (i) an artificial aortic valve that was labeled as calcium, (j) a metal artifact next to the aortic wall that was partially labeled as  calcium, (k) a false positive detection in proximity of the aorta, (l) a correctly detected calcification in LCX that was incorrectly labeled as LAD.}
	\label{fig:Examples}
\end{figure*}

\subsection{Voxel-level vs.\ lesion-level annotation}

The manual reference annotation was performed voxel-by-voxel to enable annotation of scans with poor image quality. To assess how much voxel-level annotation differs from the standard lesion-level annotation, we converted the voxel-level annotations into lesion-level annotations using 3D region growing with the standard calcium threshold (\SI{130}{HU}). Lesions were labeled using majority voting if they contained voxels with different labels. Scans with a more than five times increase in calcium volume were excluded. These were \SI{21}{\percent} of the scans with soft reconstruction kernel and \SI{64}{\percent} of the scans with sharp reconstruction kernel. In the remaining scans, the overall Agatston score increased on average by 85 in soft reconstructions and by 155 in sharp reconstructions. This clearly indicates that in low-dose CT scans, lesion-level annotation leads to an overestimation of the calcium score.

\section{Discussion} \label{sec:Discussion}

We proposed a method for automatic detection of CAC subdivided into LAD, LCX and RCA calcifications, TAC and cardiac valve calcifications in low-dose chest CT. The method is the first that detects these calcifications concurrently. The approach is based on two consecutive CNNs: The first CNN uses stacked dilated convolutions to facilitate a large receptive field, which enables identification and spatial labeling of high density voxels. The second CNN discards false positive detections of the first CNN.

For CAC and TAC detection, the method achieved a performance close to the level of interobserver agreement. The method was furthermore able to separate calcifications in the coronary arteries into LAD, LCX and RCA calcifications (\Cref{fig:Examples}\,(f)). The method as well as the observers were more successful in identification of LAD and RCA calcifications than LCX calcifications. The course of LCX is particularly difficult to follow in non-contrast scans. Hence, LCX calcifications can be difficult to differentiate from LM and LAD calcifications (\Cref{fig:Examples}\,(l)), as well as from those in the mitral valves.
In comparison to CAC and TAC, calcifications of the aortic and mitral valves were less common in our dataset. Performance of the automatic method was below the performance of CAC and TAC detection. However, this is also a difficult task for experts. The observers especially disagreed on mitral valve calcifications, which is in line with findings of previous studies \cite{vanHamersvelt2014}. The disagreement is mainly caused by confusion with LCX calcifications and the lack of soft tissue contrast in the mitral valve region. For the aortic valve, confusion with TAC was the most common cause of disagreement.

False positive detections were mostly caused by mislabeling of calcifications with respect to their location (e.g., LAD and LCX), low-dose and motion artifacts, and other calcifications such as calcified lymph nodes or calcifications in other vessels (\Cref{fig:Examples}\,(i)--(k)). False positive detections outside the heart and the aorta occurred infrequently and usually in proximity to the heart or the aorta. This demonstrates that \CNN{1} was able to implicitly learn to recognize the typical spatial context of calcifications in the image. The individual evaluation of \CNN{1} additionally showed that \CNN{2} substantially contributes to reducing false-positive detections while maintaining a high sensitivity. However, future work could aim at unifying the two networks into a single network.

False negative detections were sometimes partially misclassified lesions (\Cref{fig:Examples}\,(h)--(i)). Partial misclassification can occur because the method performs voxel classification rather than the standardly used lesion classification. Even though voxel labeling occasionally causes partial misclassification of calcifications, it enables splitting of calcifications that are contained in more than one arterial bed, such as those partly located in the aorta and partly in the coronary arteries (\Cref{fig:Examples}\,(f)). Assigning a calcification that is partially contained in the aorta to the coronary artery could affect cardiovascular risk categorization. Similarly, assigning LM calcifications to the aorta would result in missing high risk lesions. To the best of our knowledge, this is the first method enabling splitting of the calcifications according to their arterial bed.

Other methods for calcification detection often first detect a region of interest in the image. The proposed method is able to omit this step and instead searches the entire image for calcifications without the need for any preprocessing steps. Moreover, the method does not require explicit spatial features, even though these have been reported in the literature as crucial for automatic calcium scoring. The results demonstrate that a CNN with dilated convolutions is able to recognize the spatial context in three orthogonal 2D patches.

In contrast, other commonly used network architectures have various shortcomings that make them less suited for calcium scoring in low-dose chest CT.
For example, U-net\cite{Ronneberger2015} is not well suited for sparse problems because it can process 3D volumes only in smaller tiles, of which most would not contain any calcium voxels. Residual networks\cite{He2015} and the similar Densely connected networks\cite{Huang2017} use pooling to increase their receptive field, which is not compatible with the idea of purely convolutional networks. Classifying all voxels in a typical 3D chest CT volume with these networks would be inefficient and time-consuming.
HoughNets\cite{Milletari2017} are based on the idea of enforcing learned shape priors, which is useful in segmentation tasks of structures with relatively homogeneous shape. However, the shape of calcifications is rather heterogeneous, especially if scans are distorted by cardiac motion. Spatial transformer networks\cite{Jaderberg2015} address alignment issues, but chest CT scans are fairly standardized. The scanned subject typically lies on the back and the FOV of the reconstructed image is standardly configured using the outer body contour or the ribs, and the apex/base of the lungs as landmarks.

A particular strength of this paper is the large, diverse and realistic dataset of low-dose chest CT scans from the NLST that we used for training and evaluation. Even though this data is challenging due to low radiation dose, the lack of ECG-synchronization and the high diversity of image acquisition parameters, the automatic method achieved good detection performance and high agreement in risk categorization.
The separate evaluation in images reconstructed with soft and sharp filter kernels additionally demonstrates that the performance on soft reconstructions does not suffer when sharp reconstructions are added to the training data. This indicates that the networks were able to generalize to both types of reconstructions.

The high reliability of the risk categorization indicates that this method can be used for cardiovascular risk assessment in lung cancer screening. While standardized risk categories are defined for CAC scores, TAC and cardiac valve calcium scores are currently not commonly used for cardiovascular risk assessment. Automatic scoring of these calcifications enables evaluation of their predictive value using available large datasets from lung screening trials, or other screening trials with CT imaging visualizing the heart.


\section*{Acknowledgment}
\addcontentsline{toc}{section}{Acknowledgment}

We are grateful to the United States National Cancer Institute (NCI) for providing access to NCI's data collected by the National Lung Screening Trial. The statements contained in this paper are solely ours and do not represent or imply concurrence or endorsement by NCI.
To enable other researchers to request the same data and perform a comparison with the here presented results, NCI has been provided with the list of scans used in this study.



\end{document}